# Fast Evaluation of Low-Thrust Transfers via Deep Neural Networks

Yue-he Zhu[*], Ya-zhong Luo[†]

National University of Defense Technology, 410073, Changsha, People's Republic of China

**Abstract** The design of low-thrust-based multitarget interplanetary missions requires a method to quickly and accurately evaluate the low-thrust transfer between any two visiting targets. Complete evaluation of the low-thrust transfer includes not only the estimation of the optimal fuel consumption but also the judgment of transfer feasibility. In this paper, a deep neural network (DNN)-based method is proposed for quickly evaluating low-thrust transfer. An efficient database generation method is developed for obtaining both the infeasible and optimal transfers. A classification DNN and a regression DNN are trained based on the infeasible and optimal transfers to judge the transfer feasibility and estimate the optimal fuel consumption, respectively. The simulation results show that the well-trained DNNs are capable of quickly determining the transfer feasibility with a correct rate of greater than 98% and approximating the optimal transfer fuel consumption with a relative estimation error of less than 0.4%. The tests on two asteroid chains further show the superiority of the DNN-based method for application to the design of low-thrust-based multitarget interplanetary missions.

## I. Introduction

Low-thrust-based multitarget interplanetary missions to explore the solar system are of great interest to space agencies because electric propulsion has much higher high specific impulses, and the ability to visit multiple asteroids or planets in one mission obviously reduces the costs and increases the scientific rewards [1]. One of the most important steps in the design of multitarget interplanetary missions is the determination of a good flight scheme. This step is essentially to solve a global trajectory optimization problem (GTOP) that resembles the traveling salesman problem (TSP). Compared with the traditional TSP, a low-thrust-based GTOP is much more difficult to solve because of not only the much larger search space caused by the dynamic of the visiting targets but also the challenge that arises in the evaluation of the low-thrust transfer. Obtaining accurate optimal fuel consumption for a low-thrust transfer is much more time consuming and computationally difficult than computing the distance between two fixed-location cities. Millions or even billions of possible transfers are required to be evaluated in sequence optimization, and it is apparently impracticable to optimize the low-thrust trajectory for each transfer while optimizing the visiting sequence. Fast estimation of the optimal fuel consumption is necessary for the design of low-thrust-based multitarget interplanetary missions.

[*] Ph.D. Candidate, College of Aerospace Science and Engineering; zhuyuehe@nudt.edu.cn.
[†] Professor, College of Aerospace Science and Engineering; luoyz@nudt.edu.cn. Senior Member AIAA.

This study aims to develop an efficient method that can quickly evaluate a short low-thrust transfer (i.e., the transfer that circles around the central body for less than a revolution), such as the transfer cases in the seventh edition of the Global Trajectory Optimization Competition (GTOC-7) [2]. A multi-impulse transfer can always be realized as long as the fuel is sufficient. However, a low-thrust-based spacecraft may not be able to reach an expected visiting target within the limited transfer time even if there is enough fuel, and this effect means the low-thrust transfer from a visiting target to the next is not always feasible. Consequently, the ability to quickly judge the transfer feasibility is also necessary when evaluating a low-thrust transfer, and the judgment of transfer feasibility must be accomplished before estimating the optimal fuel consumption because estimating the optimal fuel consumption for an infeasible transfer makes no sense.

Several analytical methods for approximating a low-thrust transfer have been proposed, but few of them are appropriate for the general case of the transfer between orbits of arbitrary eccentricity [3-6]. The Lambert method is capable of quickly evaluating the low-thrust transfer between any two bodies and is frequently used in GTOC [7-8]. This method estimates the optimal fuel consumption of a low-thrust transfer according to the velocity increment of the corresponding Lambert (two-impulse) transfer and judges the transfer feasibility by comparing the low-thrust-accumulated velocity increment and the corresponding Lambert velocity increment. Many participants, including the champion team (Jet Propulsion Laboratory, JPL), applied this method to approximate the low-thrust transfer in GTOC-7 [2]. However, the approximating performance of the Lambert method is usually not satisfactory. A more reliable method for quickly judging the transfer feasibility and estimating the optimal fuel consumption is required for the design of the low-thrust-based multitarget interplanetary missions.

Machine learning (ML) has been rapidly developed for decades [9] and widely applied in many fields, including spacecraft trajectory optimization and prediction [10-17]. To avoid expensive evaluations of the objective function when solving

GTOPs, Ampatzis and Izzo [10] tried an ML-based model during the evolutionary optimization process and presented some preliminary but very encouraging results. These authors opened up the application of learning-based methods for spacecraft trajectory optimization. The related works that have emerged in recent years are noteworthy and can be generally divided into three types. The first type of application is to train an estimator based on a number of optimized solutions to quickly evaluate the optimal velocity increment or fuel consumption for numerous transfers without optimizing them one by one, such as the accessibility assessing for the main-belt asteroids [11-12]. The attempts by Sánchez-Sánchez [13-14] and Schiavone [15] to apply an ML-based model as an on-board representation for the optimal guidance profile can be classified as the second type. The representative work of the third type is reported by Peng and Bai [16-17] and shows that an ML-based model can also be combined with physics-based models to improve the orbit prediction accuracy by learning space environment information from large amounts of observed data. This study belongs to the first type. In fact, earlier research on approximating low-thrust transfers was presented in [18]. In this preliminary study, the superiority of applying a learning-based method to estimate the optimal low-thrust fuel consumption was verified. However, the performance was not satisfactory enough because of the inappropriate selection of the learning features and the limited approximation ability of traditional ML models. Moreover, the lack of consideration for the transfer feasibility became the largest issue. As mentioned above, the complete evaluation of a low-thrust transfer includes not only the estimation of the optimal fuel consumption but also the judgment of transfer feasibility. In essence, the judgment of transfer feasibility is a classification problem, and the estimation of the optimal fuel consumption is a regression problem. Both a regressor and a classifier are required to completely evaluate a low-thrust transfer.

A deep neural network (DNN), an important member of the ML family, is a powerful learning model referring to an artificial neural network with more than one hidden layer [19]. A DNN with an appropriate network structure and activation function is expected to have a stronger approximation ability than traditional ML models [20]. The significant achievements of AlphaGo [21] and OpenAI [22] have increasingly attracted attention on DNNs and revealed a promising prospect of DNN-based applications. Owing to its powerful approximation ability, a DNN is applied to evaluate low-thrust transfers in this paper. A classification DNN and a regression DNN are trained to judge the transfer feasibility and estimate the optimal fuel consumption, respectively. The most appropriate learning features and network scale of these DNNs (i.e., the number of nodes and hidden layers) are investigated for both the judgment of transfer feasibility and the estimation of the optimal fuel consumption. The superiority of the DNN-based method for evaluating low-thrust transfers is demonstrated by numerical simulations.

The contributions of this paper are summarized as follows:

1) It is first verified that there exists a boundary between feasible and infeasible low-thrust transfers, and the transfer feasibility can be quickly and accurately determined based on the learning method.

2) A DNN-based method is developed for quickly judging the transfer feasibility and estimating the optimal fuel consumption of low-thrust transfers, and this method is verified to be practical for application to the design of low-thrust-based multitarget interplanetary missions.

The remainder of this paper is organized as follows. Section II briefly describes the low-thrust trajectory optimization method. Section III studies the feasibility of low-thrust transfer. Section IV presents the complete process of the DNN-based method for evaluating low-thrust transfers, as well as the configuration and the training method for the classification and regression DNNs. Detailed simulations for determining the most appropriate learning features and network scales of the two learning problems and a demonstration of the superiority of the DNN-based method for evaluating low-thrust transfers are given in Section V. Conclusions are drawn in Section VI.

## II. Low-Thrust Trajectory Optimization Method

The motion of a low-thrust-based spacecraft flying around the Sun can be modeled as

$$\begin{aligned} \dot{\boldsymbol{r}} &= \boldsymbol{v} \\ \dot{\boldsymbol{v}} &= -\frac{\mu}{r^3}\boldsymbol{r} + \frac{T_{\max}}{m}\boldsymbol{u}, \\ \dot{m} &= -\frac{T_{\max} \cdot \|\boldsymbol{u}\|}{I_{sp} g_0} \end{aligned} \qquad (1)$$

where $\boldsymbol{r}$ and $\boldsymbol{v}$ are the position and velocity in the heliocentric ecliptic reference frame, respectively; $m$ is the instantaneous mass of the spacecraft; $T_{\max}$ refers to the maximal thrust magnitude; $\boldsymbol{u}$ is a control vector, where $\|\boldsymbol{u}\| \in [0,\ 1]$; and $\mu, g_0$ and $I_{sp}$ denote the gravitational parameter, the standard gravity on Earth and the specific impulse of the low-thrust engine, respectively, where $\mu = 1.32712440018\text{e}11 \text{ km}^3/\text{s}^2$ and $g_0 = 9.80665 \text{ m/s}^2$. The goal is to minimize the fuel consumption of the transfer, which can be expressed as

$$J = \min\left(\frac{T_{\max}}{I_{sp}g_0}\int_{t_0}^{t_f} u \, dt\right), \tag{2}$$

where $t_0$ and $t_f$ are the initial and final transfer times, respectively. The following constraints must be satisfied for the spacecraft:

$$\begin{aligned} \boldsymbol{r}(t_0) &= \boldsymbol{r}_{c0}, \ \boldsymbol{v}(t_0) = \boldsymbol{v}_{c0}, \ m(t_0) = m_0 \\ \boldsymbol{r}(t_f) &= \boldsymbol{r}_{tf}, \ \boldsymbol{v}(t_f) = \boldsymbol{v}_{tf}, \ m(t_f) \geq m_{dry} \end{aligned}, \tag{3}$$

where $\boldsymbol{r}_{c0}, \boldsymbol{v}_{c0}$ and $\boldsymbol{r}_{tf}, \boldsymbol{v}_{tf}$ are the initial state of the chaser (spacecraft) and the final state of the rendezvous target, respectively, and $m_0$ and $m_{dry}$ are the initial mass and dry mass of the spacecraft, respectively.

A low-thrust trajectory optimization problem is essentially an optimal control problem. Due to the small convergence radius and the sensitivity of the initial guesses, it is difficult to obtain the fuel-optimal solution directly. A homotopy-based indirect method proposed by Jiang et al. [23] is applied to overcome this issue. The energy-optimal solution is first obtained, and the fuel-optimal solution is converted from the energy-optimal solution using the homotopic approach. In this study, an improved differential evolution (DE) algorithm [24] with strong global searching ability is used to find the initial values, and a sequential quadratic programming (SQP) algorithm follows to obtain the convergent fuel-optimal solution.

# III.  Feasibility of the Low-Thrust Transfer

Due to the limitations on maneuvering ability, a low-thrust-based spacecraft is not always able to transfer to the expected visiting target within a given flight time. A low-thrust transfer between two central bodies that satisfies the constraints in Eq. (3) is defined as a feasible low-thrust transfer. One that cannot yet satisfy the constraints even if the flight trajectory is optimal (i.e., the spacecraft flies toward the target with optimal thrust direction and maximal thrust magnitude throughout the transfer process) is defined as an infeasible low-thrust transfer. Domain knowledge suggests that whether a low-thrust transfer is feasible should depend on the initial state of the departure body, the final state of the rendezvous body, the initial mass of the spacecraft and the transfer time. In this section, the influence of the above factors on the transfer feasibility is studied, and a reference feasible low-thrust transfer is used for a better comparison.

Table 1 lists the initial and final states of the reference transfer, as well as the initial mass and transfer time. $T_{max}$ and $I_{sp}$ of the spacecraft are 0.3 N and 3000 s, respectively. AU = 1.49597870691e11 m is the astronomical unit. For convenience, we apply $dr_{cf}$ and $dv_{cf}$ to describe the final state of the rendezvous body instead of the orbit elements. $dr_{cf}$ and $dv_{cf}$ are the relative position and velocity to $r_{cf}$ and $v_{cf}$, respectively, where $r_{cf}$ and $v_{cf}$ are the final state of the departure body and $dr_{cf}$ and $dv_{cf}$ are described in the Vehicle Velocity Local Horizontal (VVLH) reference frame of the departure body. If some of the factors, such as the initial mass or transfer time, are changed, the reference transfer may become infeasible. To study the influence of each factor on the transfer feasibility and illustrate the relationship between feasible and infeasible transfers more clearly, the factors (i.e., $m_0$, $\Delta T$, $dr_{cf}$, $dv_{cf}$) are analyzed individually. There is no need to further analyze the influence of the initial orbit elements of the departure body because $dr_{cf}$ and $dv_{cf}$ contain the information of both the initial and final states.

Table 1 Transfer information of the reference feasible low-thrust transfer

| Initial orbit elements | $dr_{cf}$, AU | $dv_{cf}$, km/s | $m_0$, kg | $\Delta T$, day |
|---|---|---|---|---|
| [2.5 AU, 0. 001, 0, 0, 0, 0] | [0.2, 0.2, 0.2] | [1, 1, 1] | 1500 | 300 |

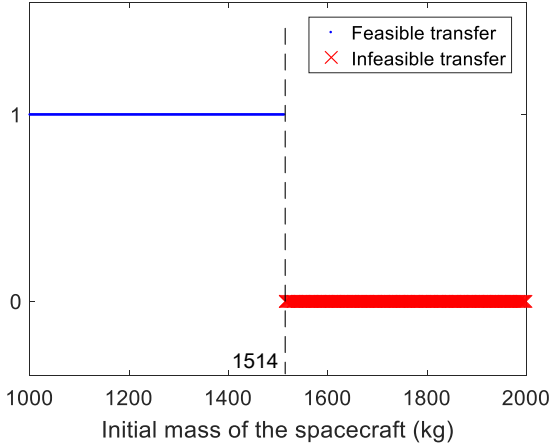 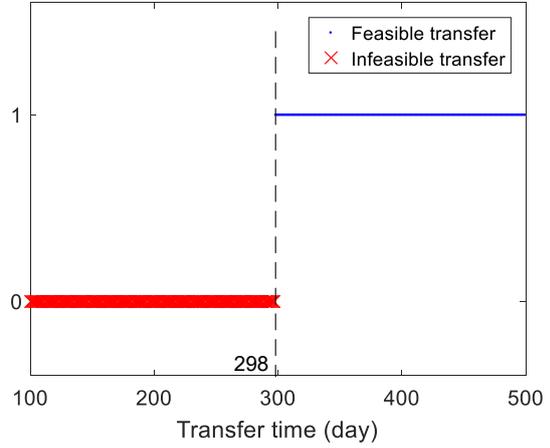

Fig. 1 Transfer feasibilities of the cases with $m_0$ ranging from 1000 to 2000 kg

Fig. 2 Transfer feasibilities of the cases with $\Delta T$ ranging from 100 to 500 days

First, the influence of the initial mass is studied. $m_0$ is incremented by 2 kg from 1000 kg to 2000 kg with all the other factors fixed. Figure 1 illustrates the transfer feasibilities of these 500 cases and shows that the transfer is infeasible if $m_0$ is larger than 1514 kg. Then, $m_0$ is set to 1500 kg, and the influence of the transfer time is studied. $\Delta T$ is incremented by one day from 100 days to 500 days, and the transfer feasibilities of these 400 cases are illustrated in Figure 2. We find that the transfer is feasible only when $\Delta T$ is longer than 298 days. The similar results in Figures 1 and 2 show that once $\Delta T$ (or $m_0$), $\boldsymbol{dr}_{cf}$ and $\boldsymbol{dv}_{cf}$ are determined, the initial mass (or the transfer time) of all the feasible transfers is limited under (or upon) a threshold.

Then, $m_0$ and $\Delta T$ are set to 1500 kg and 300 days, respectively, $\boldsymbol{dv}_{cf}$ is set to [0, 0, 0], and the influence of $\boldsymbol{dr}_{cf}$ is studied. Two conditions are considered to illustrate the relationship between feasible and infeasible transfers more clearly. The first condition is $\boldsymbol{dr}_{cf-z} = 0$, and the second condition is $\boldsymbol{dr}_{cf-x} = 0$, which means that the final positions of the transfers are all in the X-Y plane and Y-Z plane, respectively. We randomly sample 3000 points for each of the two conditions and optimize the corresponding low-thrust trajectories. Figure 3 presents the transfer feasibilities of the cases for both conditions. As shown in Figure 3, once $m_0$, $\Delta T$ and $\boldsymbol{dv}_{cf}$ are determined, the final positions of all the feasible transfers are limited in a ball-like space.

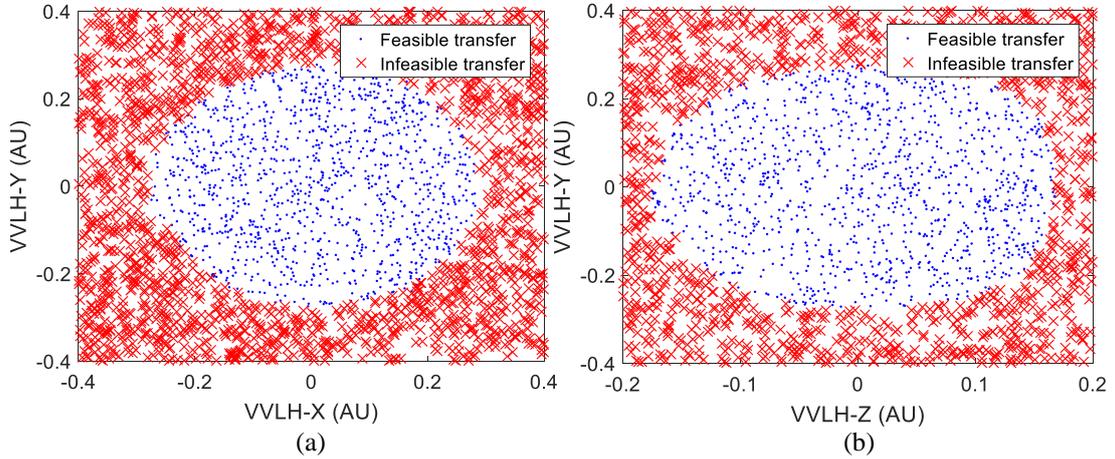

**Fig. 3 Transfer feasibilities of the cases with $\boldsymbol{dr}_{cf-z} = 0$ in (a) and $\boldsymbol{dr}_{cf-x} = 0$ in (b)**

Finally, $m_0$ and $\Delta T$ are set to 1500 kg and 300 days, respectively, $\boldsymbol{dr}_{cf}$ is set to [0, 0, 0] and the influence of $\boldsymbol{dv}_{cf}$ is studied. For better illustration, the influences of the magnitude of $\boldsymbol{dv}_{cf}$ and the direction of $\boldsymbol{dv}_{cf}$ are studied individually. We first increment $\delta$ by 0.001 from 0.7 to 1.3 to produce 600 transfer cases, where $\delta = |\boldsymbol{v}_{tf}|/|\boldsymbol{v}_{cf}|$ is the ratio of the final velocity magnitudes. The final velocity directions of these cases are the same as that of $\boldsymbol{v}_{cf}$. The transfer feasibilities of these cases are illustrated in Figure 4(a). The results indicate that once $m_0$, $\Delta T$, $\boldsymbol{dr}_{cf}$ and the direction of $\boldsymbol{dv}_{cf}$ are determined, the final velocity magnitudes of all the feasible transfers are limited in a certain range. 3000 transfer cases with the same final velocity magnitude but different directions are then randomly sampled. The transfer feasibilities of these cases are illustrated in Figure 4(b), in which the velocity vector is presented in the 3-D figure and the projection in the Y-Z plane is further affixed. The results in Figure 4(b) show that once $m_0$, $\Delta T$, $\boldsymbol{dr}_{cf}$ and the magnitude of $\boldsymbol{dv}_{cf}$ are determined, the final velocity directions of all the feasible transfers are limited in a cone beam.

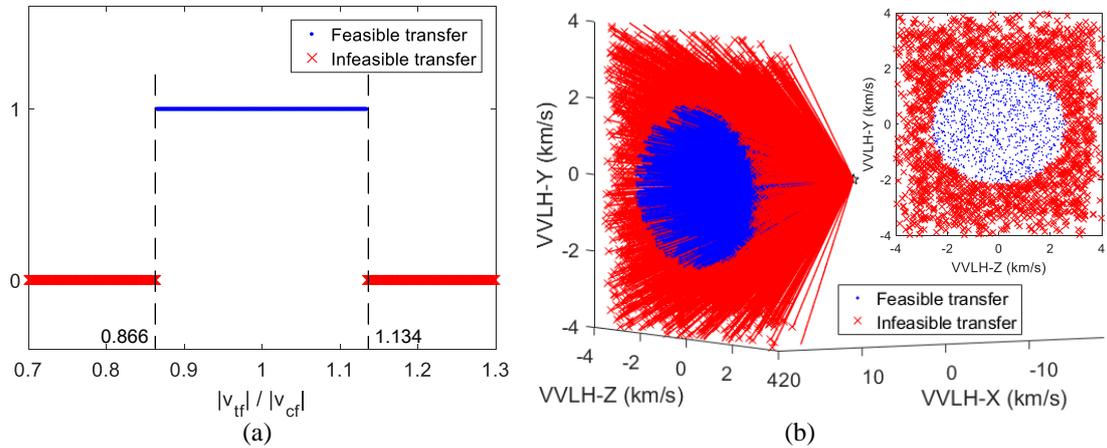

**Fig. 4 Transfer feasibilities of the cases with the same final velocity direction but different magnitudes in (a) and the same final velocity magnitude but different directions in (b)**

The above results indicate that there is a boundary between feasible and infeasible low-thrust transfers. Although it is difficult to visualize the boundary in high-dimensional space, from the above results, we can infer that the boundary is a 14-dimensional (14-D) envelope that is determined by the initial mass of the spacecraft (1-D), the flight time of the transfer (1-D), the initial state of the departure body (6-D) and the final state of the rendezvous body (6-D).

## IV. DNN-Based Method for Evaluating a Low-Thrust Transfer

Both the ability to quickly estimate the optimal fuel consumption and the ability to quickly judge the transfer feasibility are required when evaluating a low-thrust transfer. Even though it is almost impossible to analytically determine whether a low-thrust transfer is feasible and calculate the optimal fuel consumption if it is feasible, from the results in Sec. III, we know that the transfer feasibility is expected to be quickly determined with a high correct rate using a learning-based method, as long as the learning model can well approximate the boundary. The previous study [18] shows that the optimal fuel consumption is also expected to be quickly estimated with a small error using a learning-based method. A DNN is thus applied to evaluate low-thrust transfers owing to its powerful approximation ability, and a DNN-based method for judging the transfer feasibility and estimating the optimal fuel consumption is presented in this section.

### A. Implementation process

The complete process of the DNN-based method for evaluating a low-thrust transfer is divided into three steps, which are illustrated in Figure 5.

The first step is to generate the database that contains both infeasible and optimal transfers. The database should be generated according to the working conditions and parameter configurations (e.g., $T_{\max}$ and $I_{sp}$ of the spacecraft) for different problems. In fact, three types of transfers, including optimal transfers, homotopy-failed transfers and infeasible transfers, are obtained when successively generating the training data. Both optimal transfers and homotopy-failed transfers are feasible transfers. The thrust curves of three examples belonging to each of the three types are presented in Figure 6. Only the "bang-bang" controlled feasible transfers (optimal transfers), such as Example 1, and the infeasible transfers, such as Example 3, are put into the database pool. The homotopy-failed transfers, such as Example 2, are reoptimized until the corresponding optimal transfers are obtained.

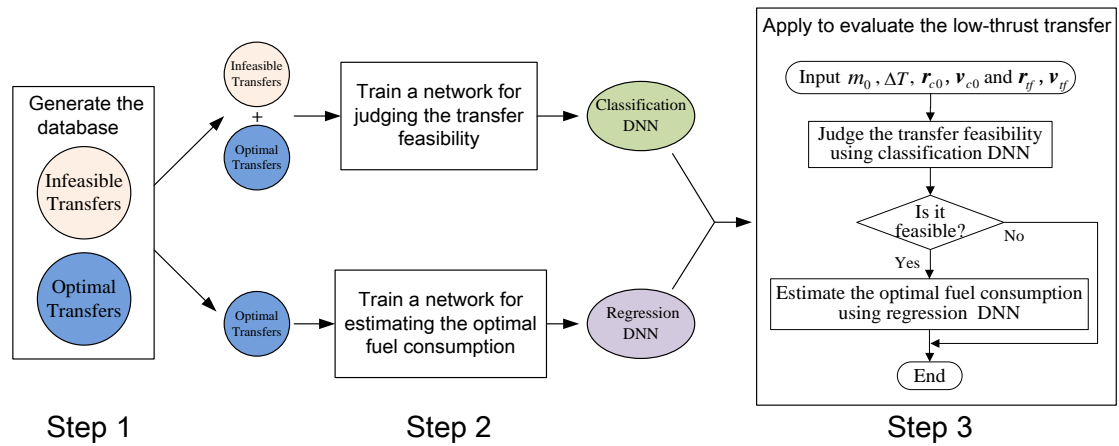

**Fig. 5 Implementation process of the DNN-based method for evaluating a low-thrust transfer**

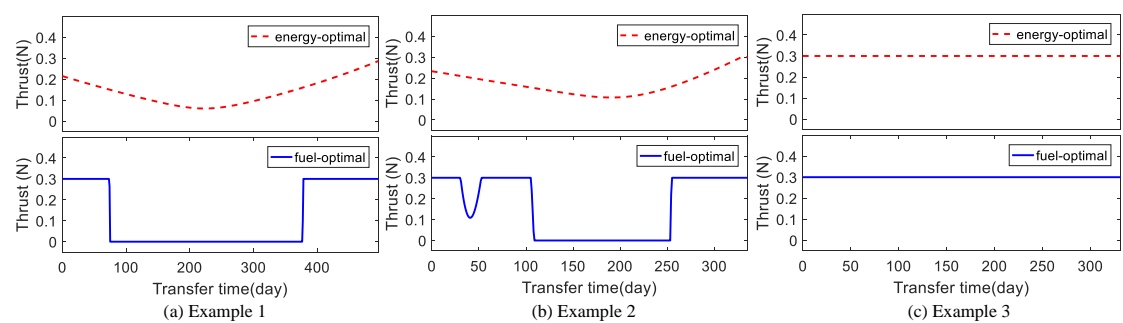

**Fig. 6 Thrust curves of the three typical examples**

The second step is to train a classification DNN and a regression DNN. Both the infeasible and optimal transfers are used to train the classification DNN, and only the optimal transfers are needed when training the regression DNN. Note that the learning features and network scales (nodes and layers) of the two DNNs are determined before taking this step. The most appropriate learning features and network scales for these two learning problems are investigated in this paper, and the results can be directly applied to similar cases when training DNNs based on the new database.

The third step is to apply the well-trained DNNs to a multitarget interplanetary mission design. The flow chart for evaluating the low-thrust transfer between any two bodies is presented in the third step and can be repeated millions or even billions of times in sequence optimization. The classification DNN is used to judge the transfer feasibility of all the input candidates, and the regression DNN focuses only on the feasible ones filtered by the classification DNN.

**B.  Database generation method**

Few feasible (optimal) transfers can be obtained if applying real-world central bodies, such as asteroids, to generate the database because the ephemeris must be considered, and most transfers cannot satisfy the phase condition. To improve the efficiency of obtaining feasible transfers and balance the proportions of the infeasible and optimal transfers in the database pool, a more efficient method is applied to generate the database.

Algorithm 1 presents the method of generating a sample (low-thrust transfer), where $Ele_{c0}$ refers to the initial orbit elements of the departure body, and $Ele_{cf}$ and $Ele_{tf}$ are the final orbit elements of the departure and rendezvous bodies, respectively. $d_1$ and $d_2$ are two parameters to determine the maxima of the final position and velocity differences. These parameters are used to control the proportions of the infeasible and optimal transfers and should be determined according to the working situation of the problem. Note that the samples generated according to this process are not the transfers between two real-world central bodies but the transfers between two virtual bodies. In fact, there is no need to use the real-world central bodies when generating the database because the DNN models trained by the transfers between virtual bodies can also be applied to evaluate the transfers between real-world bodies as long as they have the same parameter configuration.

| Algorithm 1 Pseudocode of the process to generate a training sample |
|---|
| **1:** Randomly produce $Ele_{c0}$, $m_0$ and $\Delta T$ within each range according to the working situation of the problem |
| **2:** Propagate $Ele_{c0}$ to $Ele_{cf}$ without thrusting (the propagation time is $\Delta T$) |
| **3:** Transform $Ele_{cf}$ to $r_{cf}$ and $v_{cf}$ |
| **4:** Randomly produce a position difference $dr_{cf}$ and a velocity difference $dv_{cf}$ |
| **5:** $r_{tf} = r_{cf} + dr_{cf}$, $v_{tf} = v_{cf} + dv_{cf}$ ($\|dr_{cf}\| \in [0, d_1]$, $\|dv_{cf}\| \in [0, d_2]$) |
| **6:** Transform $r_{tf}$ and $v_{tf}$ to $Ele_{tf}$ |
| **7:** Optimize the low-thrust trajectory for this sample ($m_0$, $\Delta T$, $Ele_{c0}$ and $Ele_{tf}$) |
| **8:** **if** the obtained solution is a homotopy-failed transfer |
|     return to step **7** |
|     **end if** |
| **9:** Put this sample into the database pool |

### C. DNN models and network training method

A DNN is made up of large numbers of simple, highly interconnected processing nodes. Each node takes one or more inputs from other nodes and produces an output by applying an activation function over the weighted sum of these inputs. Nodes interact using weighted connections and are arranged in layers. In this study, multilayer perceptron (MLP) is selected as the architecture for both the classification and regression DNNs. The activation of a node in MLP is determined by the summation of all the weighted inputs, which can be expressed as

$$x_j = f(\sum_{i=1}^{N} w_{ij}x_i + b_j), \tag{4}$$

where $x_j$ is the output of node $j$ in the current layer, $x_i$ is the output of node $i$ in the previous layer, $w_{ij}$ refers to the weight of the connection from node $i$ to node $j$, $b_j$ denotes the variable bias of node $j$, $N$ is the total number of nodes in the previous layer, and $f$ is the activation function. A Leaky Rectified Linear Unit (ReLU) [25] is selected as the hidden layer activation function. A sigmoid function and linear function are selected as the output-layer activation functions for the classification and regression DNNs, respectively.

Network training can be regarded as a process to adjust the weight vectors epoch by epoch, and the aim is to minimize the loss function. Based on the selection of the output layer activation function, the binary cross-entropy (BCE) function $F_c$ and the mean squared error function $F_r$ are used as the loss functions for the classification and regression DNNs, respectively, and are expressed as

$$F_c = -\frac{1}{b}\sum_{i=1}^{b}(o_m(i) \cdot \ln(o_p(i)) + (1 - o_m(i))\ln(1 - o_p(i))) \tag{5}$$

$$F_r = \frac{1}{b}\sum_{i=1}^{b}(o_p(i) - o_m(i))^2, \tag{6}$$

where $b$ is the batch size and $o_p(i)$ is the predicted output of the network. $o_m(i)$ is the transfer feasibility (0 or 1) of the input data in the training of the classification DNN and becomes the optimal fuel consumption when training the regression DNN. Cross-validation is applied in each epoch, and 90% of the data are used as training samples while the remaining 10% are used for validation. Both the classification and regression DNNs are trained until convergence with mini-batch gradient decent and a batch size of $b = 32$. The adaptive moment (Adam) [26] technique is used to optimize both the classification and regression DNNs. Keras [27] combined with TensorFlow [28] is applied to train the network, where TensorFlow is the backend of Keras.

## V. Simulations

The mission proposed in GTOC-7 [2] is exactly a low-thrust-based multitarget interplanetary mission in which the tours of the probes consist of a series of short low-thrust transfers. The demonstration of the DNN-based method for evaluating a low-thrust transfer is thus based on the mission in GTOC-7.

### A. Generating the database

Following the configuration in GTOC-7, $T_{\max}$ and $I_{sp}$ are set to 0.3 N and 3000 s, respectively. $m_0$ is limited within [800 kg, 2000 kg], and the maximum of $\Delta T$ is set to 500 days. The orbit elements of the departure and rendezvous asteroids are all within the ranges shown in Table 2. The acceptable terminal errors are set to 1e6 m and 1 m/s.

**Table 2 Ranges of the orbit elements for both the departure and rendezvous asteroids**

| $a$, AU | $e$ | $i$, deg | $\Omega$, deg | $\omega$, deg | $f$, deg |
|---|---|---|---|---|---|
| 2.0~3.0 | 0~0.4 | 0~20 | 0~360 | 0~360 | 0~360 |

The parameters $d_1$ and $d_2$ in Algorithm 1 are set to 1 AU and 10 km/s, respectively, in this case. Large numbers of transfers are obtained based on the database generation method, and approximately 40% are optimal transfers. One thousand transfers containing both the optimal and infeasible ones and another 1000 transfers containing only the optimal ones are randomly selected as the test samples for the classification and regression problems, respectively. The remaining transfers in the database pool are used as the training samples.

**B. Selection of the learning features**

An appropriate selection of the learning features is important because the lack of the relevant features and the interference of the redundant features both reduce the approximating performance [29]. A low-thrust transfer is determined by the initial state of the departure body, the final state of the rendezvous body, the initial mass of the spacecraft and the transfer time. Among these properties, the initial mass and the transfer time are two scalars that can be directly used as the learning features. The initial and final states of the transfer, however, can be expressed in different types, such as the orbit elements and the position and velocity. The possible appropriate features for judging the transfer feasibility and estimating the optimal fuel consumption are listed in Table 3, and two and three kinds of features are used to characterize the initial state of the departure body and the final state of the rendezvous body, respectively. $\Delta\theta$ and $\Delta V$, where $\Delta\theta$ is the angle between the initial and final position vectors and $\Delta V$ is the velocity increments of the corresponding Lambert transfer, are also taken into account for both the classification and regression problems to test whether they can help improve the approximating performance. The remaining mass of the spacecraft after the corresponding Lambert transfer ($m_{f-Lam}$) is further considered for the regression problem and is computed as

$$m_{f-Lam} = m_0 \cdot \exp(\frac{-\Delta V}{I_{sp} g_0}) \ . \tag{7}$$

Table 3 Possible learning features for evaluating low-thrust transfers

| Fixed features | Alternative features | | Additional features for judging the transfer feasibility | Additional features for estimating the optimal fuel consumption |
|---|---|---|---|---|
| | Initial states | Final states | | |
| $m_0$ $\Delta T$ | $Ele_{c0}$ $\boldsymbol{r}_{c0}, \boldsymbol{v}_{c0}$ | $Ele_{tf}$ $\boldsymbol{r}_{tf}, \boldsymbol{v}_{tf}$ $\boldsymbol{dr}_{cf}, \boldsymbol{dv}_{cf}$ | $\Delta\theta$ $\Delta V$ | $\Delta\theta$ $\Delta V$ $m_{f-Lam}$ |

The number of training samples is set to 5000, and a two-hidden-layer network with 30 nodes is applied to compare the approximating performance of different feature combinations for judging the transfer feasibility. Table 4 lists the results of all the tested groups. From the comparison of the first six groups, we find that Group 6 performs best with the highest judgment correct rate. This result indicates that the combination of $\boldsymbol{r}_{c0}, \boldsymbol{v}_{c0}$ and $\boldsymbol{dr}_{cf}, \boldsymbol{dv}_{cf}$ can better characterize the initial and final states of a transfer for the judgment of transfer feasibility. Based on this result, we further tested three other groups containing additional features. The results show that both $\Delta\theta$ and $\Delta V$ can contribute to the improvement of the approximating performance, and $\Delta V$ seems to have a better effect. Consequently, the features listed in Group 9 are selected as the learning features for judging the transfer feasibility.

Table 4 Correct rates of the judgment of transfer feasibility using different features

| Group | Features for judging the transfer feasibility | Correct rate |
|---|---|---|
| 1 | $m_0 + \Delta T + Ele_{c0} + Ele_{tf}$ | 0.8556 |
| 2 | $m_0 + \Delta T + Ele_{c0} + \boldsymbol{r}_{tf}, \boldsymbol{v}_{tf}$ | 0.8150 |
| 3 | $m_0 + \Delta T + Ele_{c0} + \boldsymbol{dr}_{cf}, \boldsymbol{dv}_{cf}$ | 0.8756 |
| 4 | $m_0 + \Delta T + \boldsymbol{r}_{c0}, \boldsymbol{v}_{c0} + Ele_{tf}$ | 0.8282 |
| 5 | $m_0 + \Delta T + \boldsymbol{r}_{c0}, \boldsymbol{v}_{c0} + \boldsymbol{r}_{tf}, \boldsymbol{v}_{tf}$ | 0.8170 |
| 6 | $m_0 + \Delta T + \boldsymbol{r}_{c0}, \boldsymbol{v}_{c0} + \boldsymbol{dr}_{cf}, \boldsymbol{dv}_{cf}$ | 0.8908 |
| 7 | $m_0 + \Delta T + \boldsymbol{r}_{c0}, \boldsymbol{v}_{c0} + \boldsymbol{dr}_{cf}, \boldsymbol{dv}_{cf} + \Delta\theta$ | 0.9002 |
| 8 | $m_0 + \Delta T + \boldsymbol{r}_{c0}, \boldsymbol{v}_{c0} + \boldsymbol{dr}_{cf}, \boldsymbol{dv}_{cf} + \Delta V$ | 0.9114 |
| 9 | $m_0 + \Delta T + \boldsymbol{r}_{c0}, \boldsymbol{v}_{c0} + \boldsymbol{dr}_{cf}, \boldsymbol{dv}_{cf} + \Delta\theta + \Delta V$ | **0.9208** |

Then, the number of training samples is set to 10000, and a three-hidden-layer network with 40 nodes is applied to test different feature combinations for estimating the optimal fuel consumption. For convenience, the maximum remaining mass of the spacecraft after each transfer ($m_{f-\max}$) but not the optimal fuel consumption is set to the output of the regression DNN. Table 5 lists the mean absolute errors (MAEs) of all the tested groups. From the comparison of the first six groups, we find that the results are different from those of the judgment of transfer feasibility. Not the combination of $r_{c0}, v_{c0}$ and $dr_{cf}, dv_{cf}$ but the combination of $Ele_{c0}$ and $dr_{cf}, dv_{cf}$ has the smallest MAE. This result indicates that the orbit elements can better characterize the initial state of a transfer when estimating the optimal fuel consumption. The results of the last three groups further prove that not only $\Delta\theta$ and $\Delta V$ but also $m_{f-Lam}$ can help reduce the estimation error. Consequently, all of these features are selected as the learning features for estimating the optimal fuel consumption, and they are used in the following simulations.

**Table 5 MAEs of the estimation of the optimal fuel consumption using different features**

| Groups | Features estimating the optimal fuel consumption | MAE, kg |
|---|---|---|
| 1 | $m_0 + \Delta T + Ele_{c0} + Ele_{tf}$ | 38.309 |
| 2 | $m_0 + \Delta T + Ele_{c0} + r_{tf}, v_{tf}$ | 47.624 |
| 3 | $m_0 + \Delta T + Ele_{c0} + dr_{cf}, dv_{cf}$ | 18.722 |
| 4 | $m_0 + \Delta T + r_{c0}, v_{c0} + Ele_{tf}$ | 46.551 |
| 5 | $m_0 + \Delta T + r_{c0}, v_{c0} + r_{tf}, v_{tf}$ | 46.893 |
| 6 | $m_0 + \Delta T + r_{c0}, v_{c0} + dr_{cf}, dv_{cf}$ | 22.977 |
| 7 | $m_0 + \Delta T + Ele_{c0} + dr_{cf}, dv_{cf} + \Delta\theta$ | 16.204 |
| 8 | $m_0 + \Delta T + Ele_{c0} + dr_{cf}, dv_{cf} + \Delta\theta + \Delta V$ | 12.668 |
| 9 | $m_0 + \Delta T + Ele_{c0} + dr_{cf}, dv_{cf} + \Delta\theta + \Delta V + m_{f-Lam}$ | **11.254** |

**C. Determination of the network and training data scales**

An appropriate scale of the network is necessary to avoid underfitting and overfitting. Different numbers of hidden layers and nodes are tested to determine the most appropriate network scales for both the classification and regression DNNs.

Figure 7 illustrates the correct rates of the judgment of transfer feasibility with the network scale varying from two to five hidden layers and 10 to 100 nodes. The results for the networks with more than two hidden layers show similar variation trends with the increase in the node number, where the correct rates of the judgment of transfer feasibility continue to increase before the node number reaches 40 and slowly decrease after that. The highest correct rates of the networks with two and more than three hidden layers are all worse than that of the three-hidden-layer network, and this result indicates that a network with three hidden layers and 40 nodes in each layer should be the best choice for the judgment of transfer feasibility.

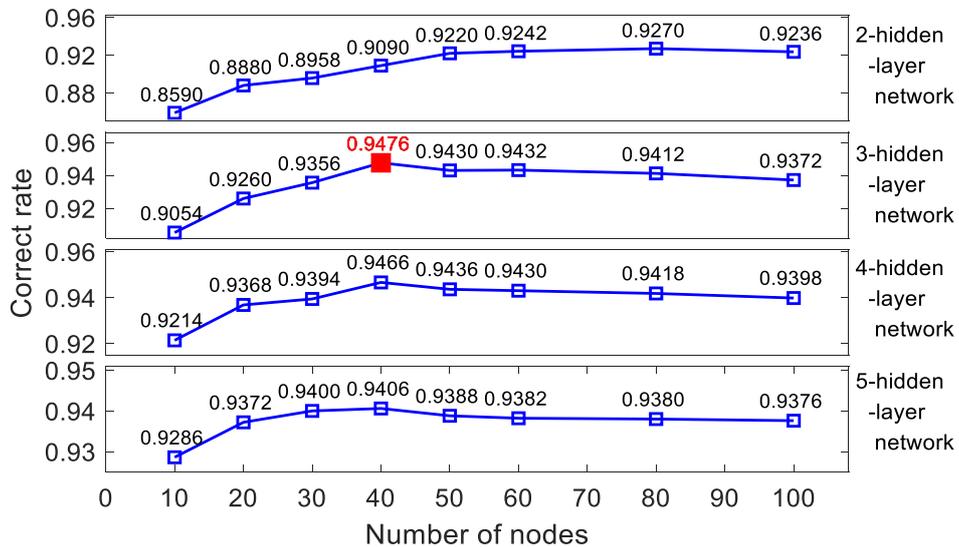

**Fig. 7 Correct rates of the judgment of transfer feasibility for networks with different numbers of hidden layers and nodes**

Based on the above result, the influence of the training data scale is further studied. Figure 8 shows that the correct rate of the judgment of transfer feasibility can be improved to as high as 98.04% if there are sufficient training samples. This result indicates that there is indeed a boundary between feasible and infeasible low-thrust transfers; otherwise, the correct rate of the judgment of transfer feasibility could not reach such a high value.

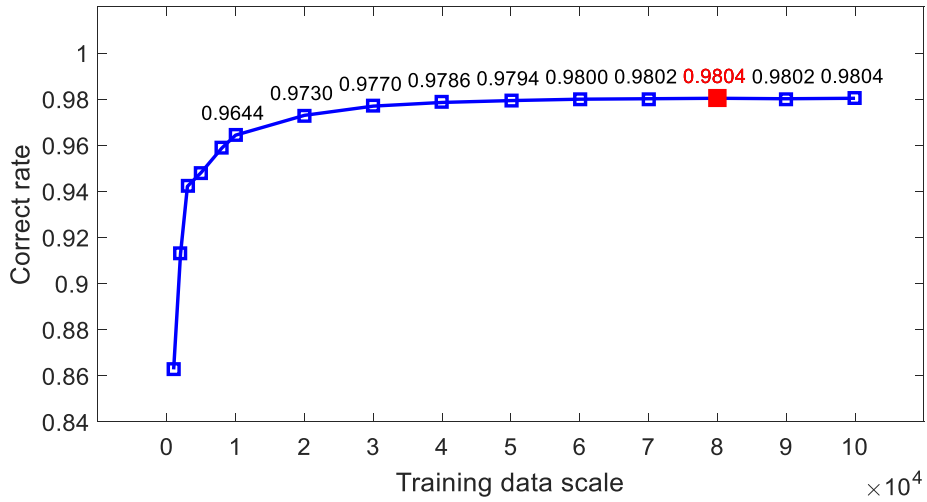

**Fig. 8 Correct rates of the judgment of transfer feasibility with different training data scales**

Figure 9 illustrates the MAEs of the estimation of the optimal fuel consumption with the network scale varying from three to five hidden layers and 20 to 100 nodes. Apparently, the network with four hidden layers and 70 nodes in each layer should be the best choice. Figure 10 shows the decrease in the MAE as the number of training samples increases from $10^4$ to $2 \times 10^5$. An amount of $1.6 \times 10^5$ samples is enough for real-world applications because further enlarging the training data scale makes no significant contribution to improvement in the approximating performance and leads to wasting time in generating the database and training the network. From the above results, we can also find that estimating the optimal fuel consumption is more difficult than judging the transfer feasibility because a larger network and training data scale are required.

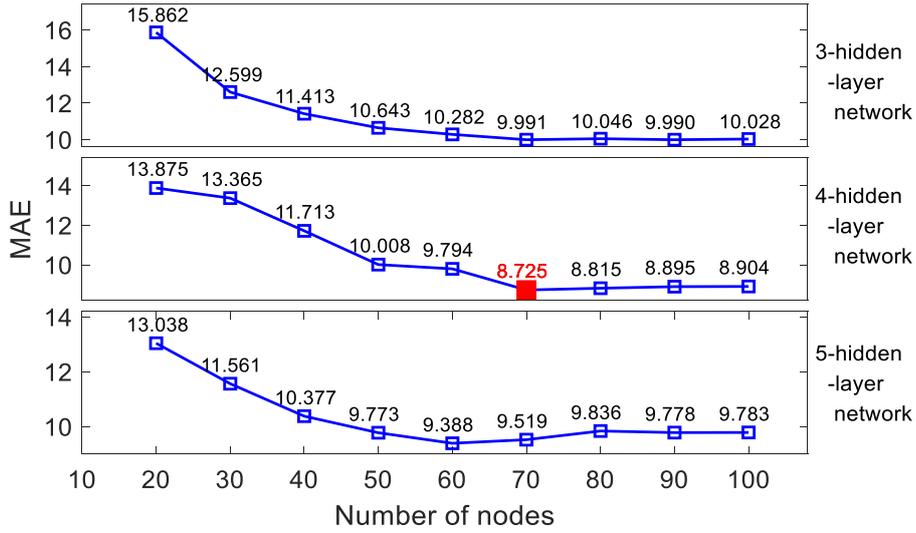

Fig. 9 MAEs of the estimation of the optimal fuel consumption with different numbers of hidden layers and nodes

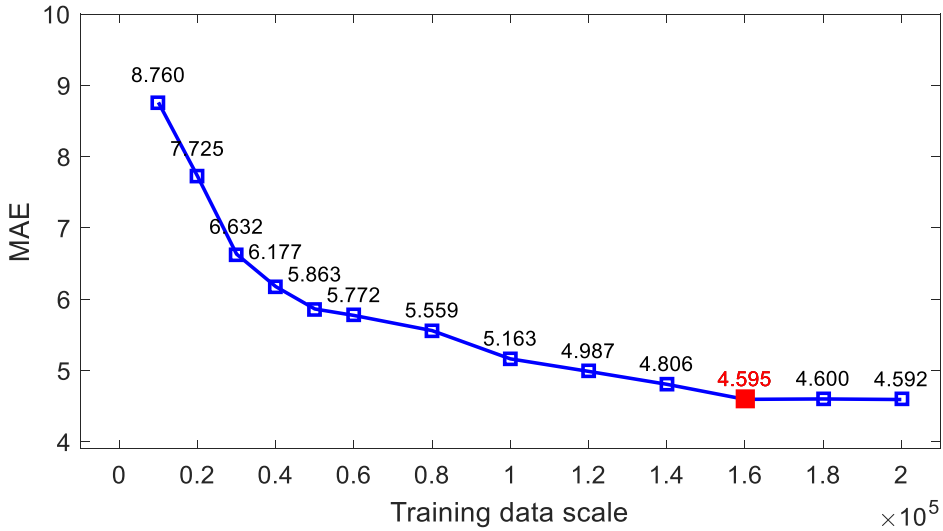

Fig. 10 MAEs of the estimation of the optimal fuel consumption with different training data scales

## D. Comparison with traditional ML-based method and Lambert method

Nine popular ML-based classifiers that are listed in Table 6, in which Groups 1~4 are single classifiers and Groups 5~9 are tree-based ensemble ones, are first tested for comparison with the classification DNN. These ML models are all trained on scikit-learn [30]. The penalty parameter of the support vector machine (SVM) is set to 100, and the maximum depth of gradient boosted classification trees (GBCT) is set to 8. The numbers of estimators in ensemble classifiers are all consistently set to 100. All of the other parameters are set to their defaults because any change in these parameters will not contribute to the improvement in the approximating performance. The correct rates of the judgment of transfer feasibility obtained by these classifiers are listed in Table 6. It can be found that the results obtained by bagging and GBCT classifiers are higher than those obtained by the other ML-based classifiers, while they are all inferior to those obtained by the classification DNN, and this result indicates that the classification DNN is more capable of judging the transfer feasibility.

**Table 6 Correct rates of the judgment of transfer feasibility using different classifiers**

| Group | Classifiers | Correct rate |
|---|---|---|
| 1 | SVM | 0.9570 |
| 2 | KNeighbors | 0.8920 |
| 3 | Gaussian process | 0.9366 |
| 4 | Decision tree | 0.9434 |
| 5 | Bagging | 0.9620 |
| 6 | AdaBoost | 0.9410 |
| 7 | GBCT | 0.9640 |
| 8 | Extremely Randomized Trees | 0.9492 |
| 9 | Random forests | 0.9558 |
| 10 | **Classification DNN** | **0.9804** |

Then, nine ML-based regressors, including four single regressors and five ensemble regressors, are tested for comparison with the regression DNN. The penalty parameter of the SVM and the maximum depth of gradient boosted regression trees (GBRT) are set to 1000 and 10, respectively. The numbers of estimators in ensemble regressors are all set to 200, and the other parameters are all set to their defaults. The simulation results listed in Table 7 show that the DNN-based method can also perform better for estimating the optimal fuel consumption than the traditional ML-based methods, and the average relative error (ARE) can be reduced to no more than 0.4%.

**Table 7 MAEs and AREs of the estimation of the optimal fuel consumption using different regressors**

| Group | Regressors | MAE | ARE |
|---|---|---|---|
| 1 | SVM | 30.455 | 2.721% |
| 2 | KNeighbors | 24.402 | 2.215% |
| 3 | Gaussian process | 25.212 | 2.380% |
| 4 | Decision tree | 26.892 | 2.437% |
| 5 | Bagging | 16.522 | 1.409% |
| 6 | AdaBoost | 34.323 | 3.082% |
| 7 | GBRT | 11.521 | 1.034% |
| 8 | Extremely Randomized Trees | 16.811 | 1.446% |
| 9 | Random forests | 12.836 | 1.157% |
| 10 | **Regression DNN** | **4.595** | **0.398%** |

The Lambert method is also compared for quickly evaluating low-thrust transfers. The Lambert method to judge the transfer feasibility is expressed as [9]

$$Feasibility = \begin{cases} 1 & \Delta V < c \cdot \Delta T \cdot T_{max} / m_0 \\ 0 & otherwise \end{cases}, \qquad (8)$$

Essentially, the transfer feasibility is judged by the comparison between the $\Delta V$ and the accumulation of the low-thrust velocity increment in $\Delta T$, and $c$ is a crucial parameter between 0 and 1 that highly influences the correct rates of judgment. We increment $c$ by 0.01 from 0 to 1 and depict the corresponding correct rates of the judgment of transfer feasibility in Figure 11. It can be seen that the correct rate obtained by the Lambert method can only reach 0.844 when $c$ is set to approximately 0.15. This result is far worse than the result obtained by the well-trained classification DNN and further shows the superiority of the DNN-based method to judge the transfer feasibility.

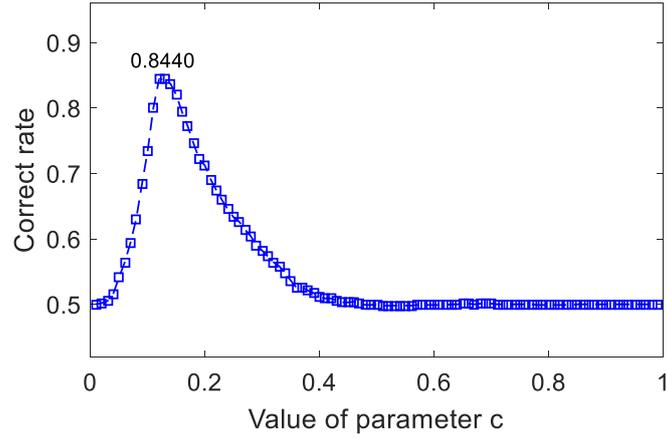

**Fig. 11 Correct rates of the judgment of transfer feasibility for the Lambert method with different values of parameter *c***

The Lambert method for estimating the optimal fuel consumption is calculated as Eq. (7). The MAE and ARE obtained by the Lambert method are 44.047 and 3.864%, respectively. These results are also far worse than the results obtained by the well-trained regression DNN. Moreover, the error distributions of the 1000 tested samples for both the DNN-based method and the Lambert method are visualized in Figure 12. The results obtained by the Lambert method show a much wider distribution, and the center of the distribution deviates from 0. This finding means that there is not only a larger random error but also a systematic error when applying the Lambert method to estimate the optimal fuel consumption. Note that $m_{f-Lam}$ is also selected as one of the learning features for estimating the optimal fuel consumption. The DNN-based method for estimating the optimal fuel consumption can consequently be seen as a technology to eliminate the systematic error and reduce the random error of the Lambert method by learning from a large number of training samples, thereby improving the estimation accuracy.

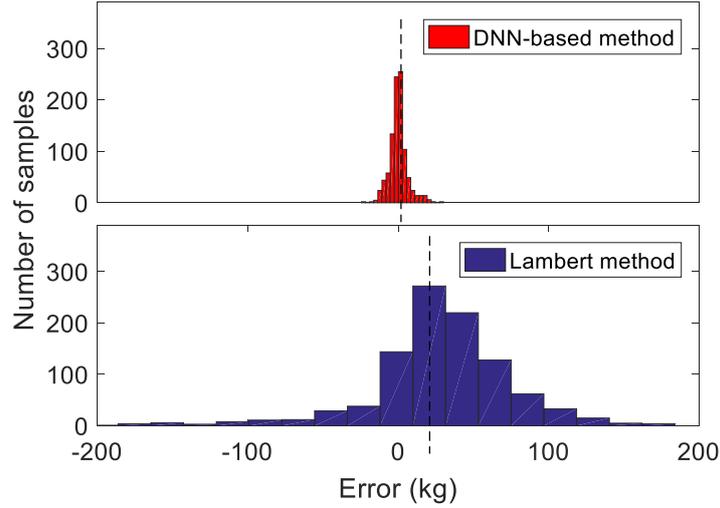

**Fig. 12 Error distributions of the tested samples obtained by the Lambert method and DNN-based method**

## E. Analysis of misjudged transfers

The judgment of transfer feasibility is a binary classification problem. Hence, there are only two kinds of misjudgment: a feasible transfer is misjudged as an infeasible one and the opposite situation. We collected all of the misjudged transfers and further checked their thrust curves. The misjudged transfers belonging to the same situation show similar thrust curves, and two examples from each of them are illustrated in Figure 13. The terminal position and velocity errors of the two misjudged transfers are also listed in Table 8.

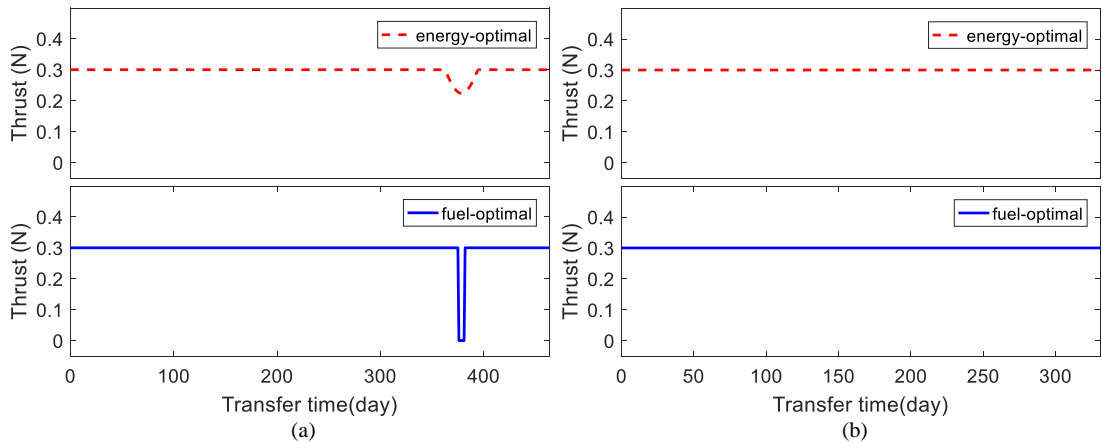

**Fig. 13 Thrust curves of misjudged transfers, where (a) is a misjudged feasible transfer and (b) is a misjudged infeasible transfer**

**Table 8 Terminal errors of the misjudged transfers**

|  | Position error, m | Velocity error, m/s |
|---|---|---|
| Misjudged feasible transfer | 4.726e2 | 0.005 |
| Misjudged infeasible transfer | 2.563e7 | 6.302 |

Figure 13(a) shows that the spacecraft has propelled with maximum thrust throughout almost the entire transfer process. Table 8 shows that the misjudged infeasible transfer only weakly violates the terminal position and velocity constraints. These results indicate that both the misjudged feasible and infeasible transfers are very close to the boundary and further show the reliability of the DNN-based method for judging the transfer feasibility.

**F. Verification on asteroid transfer chains**

To verify the effectiveness of the DNN-based method for real-world applications, the fast evaluation of successive low-thrust transfers is further studied. Two transfer chains that were achieved by the JPL team in GTOC-7 are selected as the test cases [3], where a total of 12 and 13 asteroids are contained in Chain 1 and Chain 2, respectively. The asteroid name, the rendezvous time and the optimized remaining mass after each transfer are listed in Table 9. The epoch data of the asteroids can be accessed on JPL's website [31].

**Table 9 Transfer chains in GTOC-7 obtained by JPL**

| Transfer Chain 1 | | | | Transfer Chain 2 | | | |
|---|---|---|---|---|---|---|---|
| Visiting sequence | Asteroid name | Rendezvous time, MJD | Remaining mass, kg | Visiting sequence | Asteroid name | Rendezvous time, MJD | Remaining mass, kg |
| 1 | XC77 | 61986.55 | 1878.57 | 1 | 1998 DJ10 | 61457.18 | 1995.45 |
| 2 | Stevensimpson | 62188.77 | 1782.43 | 2 | 1991 RZ8 | 61615.62 | 1899.77 |
| 3 | Clapton | 62373.71 | 1675.42 | 3 | Texereau | 61835.57 | 1760.68 |
| 4 | 1999 DS1 | 62512.13 | 1593.82 | 4 | 1998 FA57 | 62105.23 | 1596.86 |
| 5 | Roswitha | 62701.71 | 1453.22 | 5 | 1999 JE122 | 62248.66 | 1509.41 |

| 6 | Marci | 62851.55 | 1347.62 | 6 | 1991 RB9 | 62456.41 | 1404.55 |
| --- | --- | --- | --- | --- | --- | --- | --- |
| 7 | 1999 XH12 | 62958.15 | 1289.91 | 7 | 1996 XU25 | 62654.85 | 1334.8 |
| 8 | Fangfen | 63138.64 | 1185.75 | 8 | Vasifedoseev | 62798.56 | 1249.27 |
| 9 | T-2 | 63431.24 | 1006.54 | 9 | 1991 PP11 | 62941.47 | 1149.79 |
| 10 | Hukeller | 63547.71 | 948.29 | 10 | 1999 JL91 | 63078.66 | 1092.87 |
| 11 | 1991 SV | 63652.73 | 888.72 | 11 | 1998 FO47 | 63337.72 | 976.93 |
| 12 | 2000 JQ86 | 61986.55 | 847.52 | 12 | Aruna | 63491.63 | 902.62 |
| | | | | 13 | Erasmus | 61457.18 | 827.71 |

The 11 transfers in Chain 1 and 12 transfers in Chain 2 are first checked by the well-trained classification DNN, and the results show that all are feasible transfers. The remaining masses of the two chains estimated by the well-trained regression DNN are illustrated in Figures 14 and 15. The results obtained by the Lambert method are also presented for comparison. Note that the next remaining mass is estimated according to the last estimated mass but is not the true value after the first transfer because the true value is unknown without optimization.

It can be seen from Table 9 and Figures 14 and 15 that the estimation error of the Lambert method keeps increasing transfer by transfer and finally reaches an extreme value of close to 200 kg. The systematic error shown in Figure 12 causes the accumulation of the estimation error and results in a larger and larger deviation between the estimated remaining mass and the true mass. Such an estimation accuracy is unacceptable for real-world applications because there is a risk of losing the best sequence and misjudging the maximum number of accessible asteroids during sequence optimization. The dry mass of the probe is 800 kg in GTOC-7. One or even two more asteroids can be added to the tail of both Chain 1 and Chain 2 according to the estimation result of the Lambert method, while they are actually inaccessible when computing the true optimal fuel consumption.

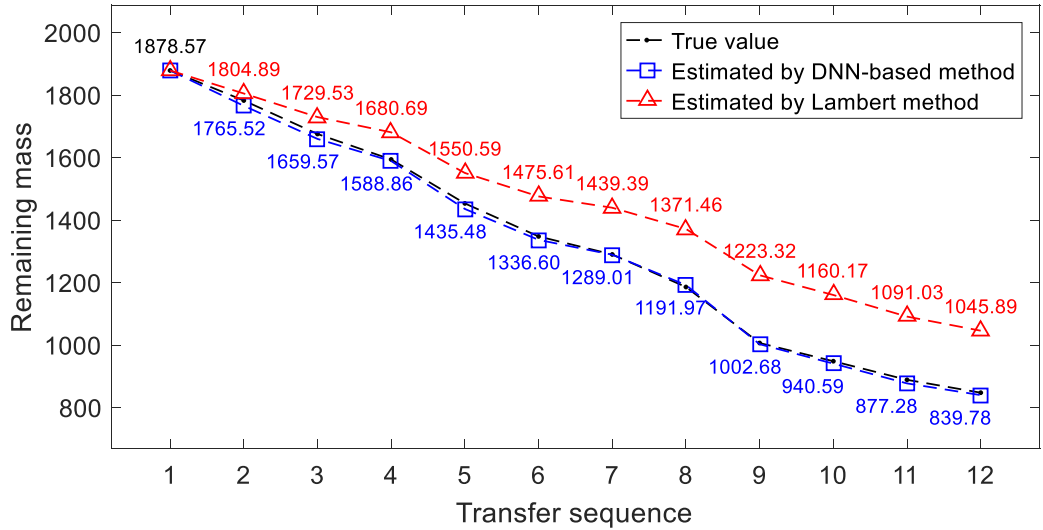

Fig. 14 True and estimated remaining mass of Chain 1

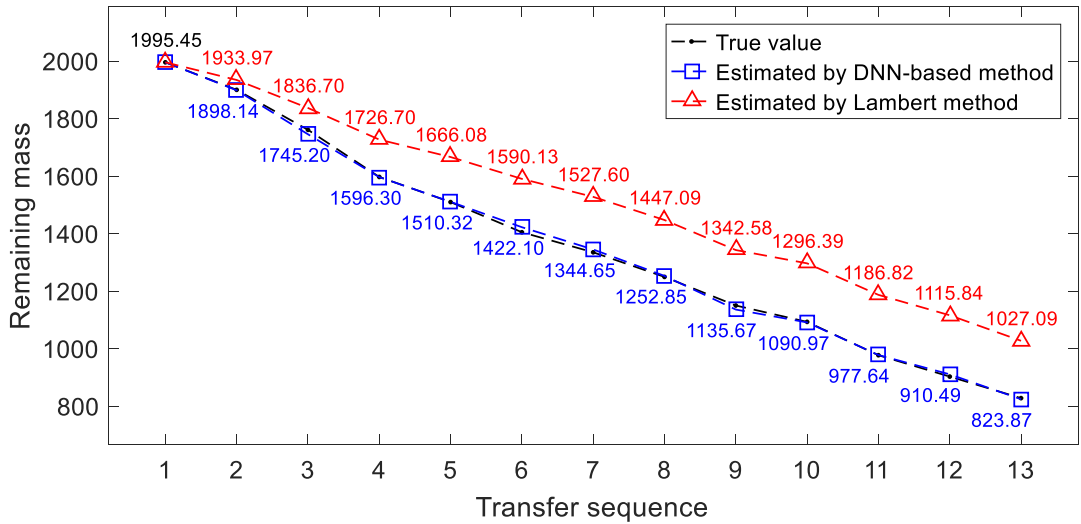

Fig. 15 True and estimated remaining mass of Chain 2

Without the interference of the systematic error and because of the offset of all the random error, the remaining mass estimated by the DNN-based method can fit the true value very well for both chains and obtain the final results with very small errors (approximately 7 kg and 3 kg for Chain 1 and Chain 2, respectively). The comparisons in these two cases better show the significant advantage of the DNN-based method for estimating the optimal fuel consumption, especially for successive low-thrust transfers.

## VI. Conclusions

Fast evaluation of low-thrust transfers is studied in this paper. The feasibility of low-thrust transfers is first analyzed, showing that a boundary exists between the feasible and infeasible transfers, and the transfer feasibility is expected to be quickly determined by the learning method. A DNN-based method for quickly evaluating low-thrust transfers is proposed, and a classification DNN and a regression DNN are required to judge the transfer feasibility and estimate the optimal fuel consumption, respectively. The implementation process of the DNN-based method as well as the methods for generating the database and training the classification and regression DNNs are presented. The most appropriate learning features and network scales are determined for judging the transfer feasibility and estimating the optimal fuel consumption. The superiority and reliability of the DNN-based method for quickly evaluating low-thrust transfers are shown by comparison with popular ML-based methods and the Lambert method. The case study on two asteroid transfer chains further reveals the advantage of the DNN-based method for estimating the optimal fuel consumption for successive low-thrust transfers.

## Acknowledgments

This work was supported by the National Natural Science Foundation of China (11572345).